# Two Questions about Data-Oriented Parsing [*]

Rens Bod

University of Amsterdam
Department of Computational Linguistics
Spuistraat 134, NL-1012 VB Amsterdam
Rens.Bod@let.uva.nl

**Abstract**

In this paper I present ongoing work on the data-oriented parsing (DOP) model. In previous work, DOP was tested on a *cleaned-up* set of analyzed *part-of-speech strings* from the Penn Treebank, achieving excellent test results. This left, however, two important questions unanswered: (1) how does DOP perform if tested on *unedited* data, and (2) how can DOP be used for parsing *word strings* that contain unknown words? This paper addresses these questions. We show that parse results on unedited data are worse than on cleaned-up data, although still very competitive if compared to other models. As to the parsing of word strings, we show that the hardness of the problem does not so much depend on unknown words, but on previously unseen lexical categories of known words. We give a novel method for parsing these words by estimating the probabilities of unknown subtrees. The method is of general interest since it shows that good performance can be obtained without the use of a part-of-speech tagger. To the best of our knowledge, our method outperforms other statistical parsers tested on Penn Treebank word strings.

## 1    Introduction

The Data-Oriented Parsing (DOP) method suggested in Scha (1990) and developed in Bod (1992-1995) is a probabilistic parsing strategy which does not single out a narrowly predefined set of structures as the statistically significant ones. It accomplishes this by maintaining a large corpus of analyses of previously occurring utterances. New input is parsed by combining tree-fragments from the corpus; the frequencies of these fragments are used to estimate which analysis is the most probable one.

In previous work, we tested the DOP method on a *cleaned-up* set of analyzed *part-of-speech strings* from the Penn Treebank (Marcus et al., 1993), achieving excellent test results (Bod, 1993a,b). This left, however, two important questions unanswered: (1) how does DOP perform if tested on *unedited* data, and (2), how can DOP be used for parsing *word strings* that contain unknown words?

This paper addresses these questions. The rest of it is divided into three parts. In section 2 we give a short resume of the DOP method. In section 3 we address the first question: how does DOP perform on unedited data? In section 4 we deal with the question how DOP can be used for parsing word strings that contain unknown words. This second question turns out to be the actual focus of the article, while the answer to the first question serves as a baseline.

---

[*] This work was partially supported by the Netherlands Organization for Scientific Research (NWO).

## 2   Resume of Data-Oriented Parsing

Following Bod (1995a), a Data-Oriented Parsing model can be characterized by (1) a definition of a formal *representation for utterance analyses*, (2) a definition of the *fragments* of the utterance analyses that may be used as units in constructing an analysis of a new utterance, (3) a definition of the *operations* that may be used in combining fragments, and (4) a definition of the way in which the *probability of an analysis* of a new utterance is computed on the basis of occurrence-frequencies of the fragments in the corpus. In Bod (1992, 1993a), a first instantiation of this model is given, called DOP1, which uses (1) labelled trees for the utterance analyses, (2) subtrees for the fragments, (3) node-substitution for combining subtrees, and (4), the sum of the probabilities of all distinct ways of generating an analysis as a definition of the probability of that analysis.

An example may illustrate DOP1. Consider the imaginary, extremely simple corpus in figure 1 which consists of only two trees.

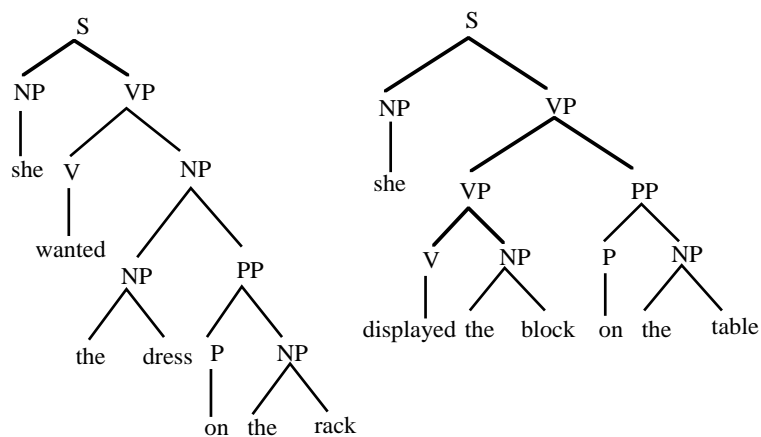

Figure 1. Example corpus of two trees.

DOP1 uses *substitution* for the combination of subtrees. The combination of subtree *t* and subtree *u*, written as $t \circ u$, yields a copy of *t* in which its leftmost nonterminal leaf node has been identified with the root node of *u* (i.e., *u* is *substituted* on the leftmost nonterminal leaf node of *t*). For reasons of simplicity we will write in the following $(t \circ u) \circ v$ as $t \circ u \circ v$. Now the (ambiguous) sentence *"She displayed the dress on the table"* can be parsed in many ways by combining subtrees from the corpus. For instance:

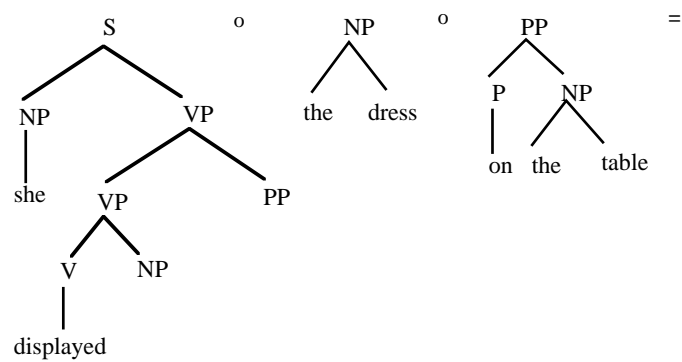

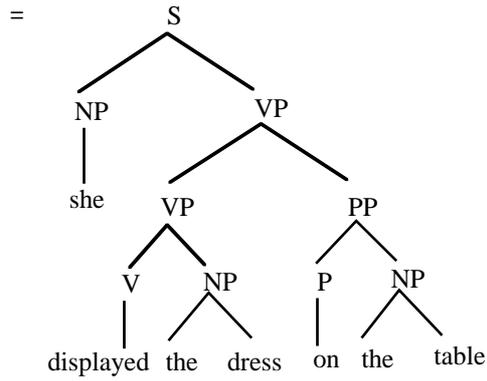

Figure 2. Derivation and parse tree for *"She displayed the dress on the table"*

As the reader may easily ascertain, a different derivation may yield a different parse tree. However, a different derivation may also very well yield the same parse tree; for instance:

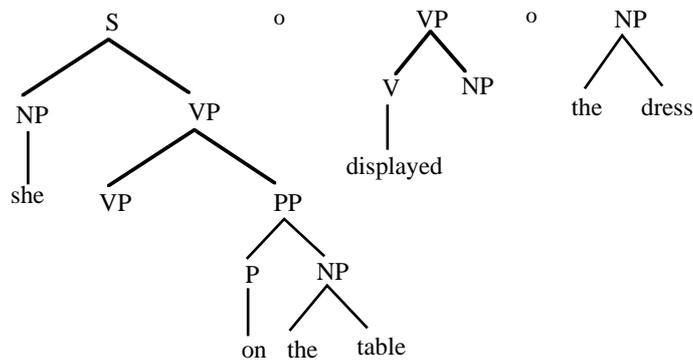

Figure 3. Different derivation generating the same parse tree for *"She displayed the dress on the table"*

or

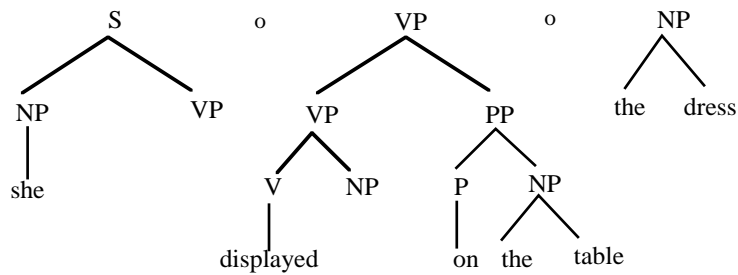

Figure 4. Another derivation generating the same tree for *"She displayed the dress on the table"*

Thus, one parse tree can be generated by many derivations involving different corpus-subtrees. DOP1 estimates the probability of substituting a subtree $t_i$ on a specific node as the probability of selecting $t_i$ among all subtrees in the corpus that could be substituted on that node. This probability can be estimated as the number of occurrences of a subtree $t_i$, divided by the total number of occurrences of subtrees $t$ with the same root node label as $t_i$: $\#(t_i) / \#(t : \text{root}(t) = \text{root}(t_i))$. The probability of a derivation $t_1 \circ ... \circ t_n$ can be computed as the product of the probabilities of the substitutions that it involves: $\Pi_i \#(t_i) / \#(t : \text{root}(t) = \text{root}(t_i))$. The probability of a parse tree is equal to the probability that any of its derivations occurs, which is the sum of the probabilities of all derivations of that parse tree. If a parse tree has $k$ derivations: $(t_{11} \circ ... \circ t_{1j} \circ ...), ..., (t_{i1} \circ ... \circ t_{ij} \circ ...), ..., (t_{k1} \circ ... \circ t_{kj} \circ ...)$, its probability can be written as: $\Sigma_i \Pi_j \#(t_{ij}) / \#(t : \text{root}(t) = \text{root}(t_{ij}))$.

Bod (1992, 1993a) shows that conventional context-free parsing techniques can be used in creating a parse forest for a sentence in DOP1. To select the most probable parse from a forest, Bod (1993-95) and Rajman (1995a,b) give Monte Carlo approximation algorithms. Sima'an (1995) gives an efficient polynomial algorithm for selecting the parse corresponding to the most probable derivation. In Goodman (1996), an efficient parsing strategy is given that maximizes the expected number of correct constituents.[1] The DOP1 model, and some variations of it, have been tested by Bod (1993-1995), Sima'an (1995-1996), Sekine & Grishman (1995), Goodman (1996), and Charniak (1996).

## 3  How does DOP perform on unedited data?

Our first question is concerned with the performance of DOP1 on unedited data. To deal with this question, we use ATIS p-o-s trees as found in the Penn Treebank (Marcus et al., 1993). This paper contains the first published results with DOP1 on unedited ATIS data.[2] Although Sima'an (1996) and Goodman (1996) also report experiments on unedited ATIS trees, their results do not refer to the most probable parse but to the most probable derivation and the maximum constituents parse respectively. We will not deal here with the algorithms for calculating the most probable parse of a sentence. These have been extensively described in Bod (1993-1995) and Rajman (1995a,b).

### 3.1  Experiments with the most probable parse

It is one of the most essential features of the DOP approach, that arbitrarily large subtrees are taken into consideration to estimate the probability of a parse. In order to test the usefulness of this feature, we performed different experiments constraining the *depth* of the subtrees. The depth of a tree is defined as the length of its longest path. The following table shows the results of seven experiments for different maximum depths of the subtrees. These were obtained by dividing the ATIS trees at random into 500 training set trees and 100 test set trees. The training set trees were converted into subtrees together with their substitution probabilities. The part-of-speech sequences from the test set served as input sentences that were parsed and disambiguated using the subtrees from the training set. The *parse accuracy* is defined as the percentage of test sentences for which the most probable parse *exactly matches* with the test set parse.

---

[1] Goodman's approach is different from Bod (1993-1995) and Sima'an (1995) in that it returns best parses that cannot be generated by the DOP1 model (see Bod, 1996 for a reply to Goodman's paper).

[2] Bod (1995b) also reports on experiments with unedited data, but the book in which that paper is included has not yet appeared.

| depth of subtrees | parse accuracy |
|:---:|:---:|
| 1 | 27 % |
| ≤2 | 45 % |
| ≤3 | 56 % |
| ≤4 | 59 % |
| ≤5 | 61 % |
| ≤6 | 62 % |
| unbounded | 64 % |

Table 3.1. Parse accuracies for DOP1 for different maximum subtree depths.

The table shows a considerable increase in parse accuracy when enlarging the maximum depth of the subtrees from 1 to 2. The accuracy keeps increasing, at a slower rate, when the depth is enlarged further. The highest accuracy of 64% is achieved by using all subtrees from the training set. If once-occurring subtrees were ignored, the maximum parse accuracy decreased to 60%. This shows the importance of taking all subtrees from the training set.

However, the accuracy of 64% is disappointingly bad when compared to exact match results reported on *clean* ATIS data: 96% in Bod (1995a), 90% to 95% in Carter & Rayner (1994). An explanation may be that the exact match metric is very sensitive to annotation errors. Since the raw Penn Treebank data contains many inconsistencies in its annotations (cf. Ratnaparkhi, 1996), a single inconsistency in a test set tree will very likely yield a zero percent parse accuracy for the particular test set sentence. Thus, we should raise the question as to whether the exact match is an interesting metric for parsing with inconsistent data.[3]

Accuracy metrics that are less sensitive to annotations errors are the so-called *bracketing accuracy* (the percentage of the brackets of the most probable parses that do not cross the brackets in the test set parses), and the *sentence accuracy* (the percentage of the most probable parses in which *no* brackets cross the brackets in the test set parses). We also calculated the accuracies according to these metrics for DOP1. To increase the reliability of our results, we performed experiments with 8 different random divisions of ATIS into training sets of 500 and test sets of 100 trees. The following table shows the means of the results for the three accuracy metrics with their standard deviations.

| Accuracy metric | Mean | StdDev |
|:---|:---:|:---:|
| Parse accuracy | 62.3 % | 3.4 % |
| Sentence accuracy | 72.8 % | 4.0 % |
| Bracketing accuracy | 94.1 % | 1.8 % |

Table 3.2. Mean accuracies for DOP1 for 8 different training-test sets from ATIS

---

[3] This question only refers to *syntactic* parsing. *Semantic* parsing turns out to be much more robust to annotation errors. In Bod, Bonnema & Scha (1996), it is shown that on semantically enriched ATIS trees, 88% of the test sentences obtain the correct semantic interpretation, while only 62% obtain a fully correct syntactic structure.

## 3.2    Comparison with other systems

Above results now allow us to compare the accuracy of DOP1 with other systems tested on unedited ATIS data. The system developed by (Pereira & Schabes 1992), where use is made of an iterative reestimation algorithm derived from the well-known inside-outside algorithm (Baker, 1979), obtains 90.4% bracketing accuracy. This is lower than our 94.1%. Pereira and Schabes do not report the sentence accuracy nor the parse accuracy of their system. The system developed by (Brill, 1993) uses a transformation-based learning algorithm. He reports 91.1% bracketing accuracy and 60% sentence accuracy on the ATIS. Although Brill does not report the parse accuracy of his system, we can derive that DOP1 does better on the bracketing accuracy and the sentence accuracy. Finally, the system of (de Marcken, 1995), who incorporates mutual information between phrase heads, obtains maximally 92.0% bracketing accuracy.

Thus, we derive that on unedited ATIS data, DOP obtains very competitive results, if not better results than other systems. This is remarkable, since DOP is *not* trained: it reads the rules (or subtrees) *directly* from hand-parsed sentences in a treebank, and calculates the probability of a new tree on the basis of *raw* subtree-frequencies in the corpus. Thus, one can seriously put into question the merits of sophisticated training and learning algorithms.

We should of course ask whether the relative succes of the DOP approach only holds for such limited domains as ATIS, or whether the approach can be effectively applied to larger corpora such as the Wall Street Journal (WSJ) corpus? Although we have not yet accomplished experiments on the WSJ, it may be interesting to report an experiment accomplished by Charniak (1996). Charniak applies the DOP approach to p-o-s strings from Penn's WSJ. Although he only uses corpus-subtrees smaller than depth 2 (which in our experience constitutes a less-than-optimal version of the DOP method -- see table 3.1), Charniak applies exactly the same statistics without ignoring low-count events. He reports that his program "outperforms all other non-word-based statistical parsers/grammars on this corpus". We conjecture that his results will be even better if larger subtrees are taken into account.

## 4    Can DOP be used for parsing word strings?

The second question mentioned in the introduction concerns the problem of parsing word strings. Since DOP1 only uses subtrees that are literally found in a training set, it cannot adequately parse or disambiguate sentences with unknown words. In this section, we investigate what is involved in extending DOP1 in order to parse sentences that possibly contain unknown words. The problem is of general importance, since practically all methods seem to have accepted the use of a two step approach: first tag the words by a part-of-speech tagger, then parse the tags (with or without the words) by a stochastic parser. We strongly believe that such a two step approach is not optimal (see section 4.3.3), and we therefore want to cope with word parsing by skipping the p-o-s tagging step.

### 4.1    The model DOP2: the partial parse method

In order to get a feeling of the problems emerging from word parsing, we propose as a very first tentative solution the model DOP2. DOP2 is a very simple extension of DOP1: assign all lexical categories to each unknown word, and select the most probable parse among the parses of all resulting "sentences" by means of DOP1. Thus, unknown words are assigned a lexical category such that their

"surrounding" partial parse has maximal probability. We shall refer to this method as the *partial parse method*.

The computational aspects of DOP2 are straightforward: we can basically employ the same algorithms as developed for DOP1 (Bod, 1995a). We only need to establish the unknown words of an input sentence and label them with all lexical categories. Remember that for input sentences *without* unknown words, DOP2 is identical to DOP1.

**4.2     Experiments with DOP2: the problem of unknown-category words**

In our experiments with DOP2, we used the same initial division of the ATIS corpus as in section 3.1 into a training set of 500 trees and a test set of 100 trees, but now the trees were *not* stripped of their words. For time-cost reasons, no experiments were performed with subtrees larger than depth 3. The following table gives the results of these experiments (for subtree-depth $\leq$ 3). We represented respectively the parse accuracy of the test sentences that contained at least one unknown word, the parse accuracy of the test sentences without unknown words, and finally, the parse accuracy of all test sentences together (we will come back to the other accuracy metrics later).

| test sentences | parse accuracy |
|---|---|
| with unknown words | 20% |
| with only known words | 33% |
| all test sentences | 24% |

Table 4.1. Parse accuracy for word strings from the ATIS corpus by DOP2

The table shows that the results of the partial parse method are disappointingly bad. For the sentences with unknown words, only 20% are parsed correctly. However, if plain DOP1 were used, the accuracy would have been 0% for these sentences. If we look at the sentences with only known words (where DOP2 is equivalent to DOP1), we see that the parse accuracy of 33% is higher than for sentences with unknown words. However, it remains far behind the 56% parse accuracy of DOP1 for part-of-speech strings (for the same subtree depth; see table 3.1). Word parsing is obviously a much more difficult task than part-of-speech parsing, even if all words are known.

Looking more carefully to the parse results of test sentences with only known words, we discover a striking result: for many of these sentences *no* parse could be generated at all, not because a word was unknown, but because an ambiguous word required a lexical category which it didn't have in the training set. We will call these words *unknown-category words*.

One might argue that the problem of unknown-category words is due to the tiny size of the ATIS corpus. However, no corpus of any size will ever contain all possible uses of all possible words. Even the extension with a dictionary does not solve the problem. There will be domain-specific words and word senses, abbreviations, proper nouns etc. that are not found in a dictionary. It remains therefore important to study how to deal with unknown words and unknown-category words in a statistically adequate way.

## 4.3 The model DOP3: a corpus as a sample of a larger population

We have seen that the partial parse method, employed by DOP2, yields very poor predictions for the correct parse of a sentence with (an) unknown word(s). For sentences with unknown-category words, the method appeared to be completely inadequate. A reason for these shortcomings may be the statistical inadequacy of DOP2: it does not allow for the computation of the probability of a parse containing one or more unknown words. In this section, we study what is involved in creating an extension of DOP1 which can compute probabilities of parses containing unknown (-category) words. In order to create such an extension, we need a method that estimates the probabilities of unknown *subtrees*.

### 4.3.1 The problem of unknown subtrees

By an unknown subtree we mean a subtree which does not occur in the training set, but which may show up in an additional sample of trees (like the test set). We will restrict ourselves to subtrees whose unknownness depends only on unknown terminals. We assume that there are no unknown subtrees that depend on an unknown syntactic structure. Thus, if there is an unknown subtree in the test set, then there is a subtree in the training set which differs from the unknown subtree only in some of its terminals. In general, this assumption is wrong, but for the ATIS domain it may not be unreasonable. In principle, the assumption that only the terminals in unknown subtrees may be unknown can be abolished, but this would lead to a space of possible subtrees which is computationally intractable.

Even with the current restriction, the problem is far from trivial. Two main questions are:

1. How can we derive unknown subtrees?
2. How can we estimate the probabilities of unknown subtrees?

As to the first question, we are not able to generate the space of unknown subtrees in advance, as we do not know the unknown terminals in advance. But since we assume that all syntactic structures have been seen, we can derive the unknown subtrees that are needed for parsing a certain input sentence, by allowing the unknown words and unknown-category words of the sentence to *mismatch* with the lexical terminals of the training set subtrees. The result of a mismatch between a subtree and one or more unknown (-category) words is a subtree in which the terminals are replaced by the words with which it mismatched. In other words, the subtree-terminals are treated as if they are wildcards. As a result, we may get subtrees in the parse forest that do not occur in the training set.

The mismatch-method has one bottleneck: the unknown words and unknown-category words of a sentence need to be known before parsing can start. It is easy to establish the *unknown* words of a sentence, but it is unclear how to establish the *unknown-category* words. Since every word is a potential unknown-category word (even closed-class words, if a small corpus is used), we ideally need to treat all words of a sentence as possible unknown-category words. Thus, any subtree-terminal is allowed to mismatch with any word of the input sentence. (In our experiments, however, we need to limit the potential unknown-category words as much as possible; cf. §4.5.)

How can we estimate the probability of a subtree which appears as a result of the mismatch-method in the parse forest, but not in the training set? It is evident that we cannot assume, as we did in DOP1, that the space of training set subtrees represents the total population of subtrees, since this would lead to a zero probability for any unknown subtree. We therefore pursue an alternative approach, and treat the space of subtrees as a sample of a larger population. We believe that such an approach is also reasonable from a cognitive point of view.

An unknown subtree which has a zero probability in a sample may have a non-zero probability in the total population. Moreover, also known subtrees may have population probabilities that differ from their sample probabilities. The problem is how to estimate the population probability of a subtree on the basis of the observed sample. Much work in statistics is concerned with the fitting of particular distributions to sample data, with or without motivating why these distributions might be expected to be suitable. A method which is largely independent of the distributions of population probabilities is the so-called *Good-Turing* method (Good, 1953). It only assumes that the sample is obtained at random from the total population. We will propose this method for estimating the probabilities of unknown subtrees, as well as of known subtrees. We briefly describe the method in the following section and then go into the problem of applying Good-Turing to subtrees (these sections heavily rely on Church & Gale, 1991). Although more sophisticated methods exist as well, we will stick to Good-Turing for the scope of this paper.

### 4.3.2   Good-Turing: estimating the population probabilities of (un)seen types

The Good-Turing method, (Good, 1953), estimates the probability of a type *t* by adjusting its observed sample frequency *f(t)*. To do that, Good-Turing uses an additional notion, represented by $N_r$, which is defined as the number of types which are instantiated by *r* tokens in an observed sample: $N_r = \#(\{ t \mid f(t) = r \})$. Thus, $N_r$ is the frequency of frequency *r*. The Good-Turing estimator computes for every frequency *r* an adjusted frequency *r\** as

$$r^* = (r+1) \frac{N_{r+1}}{N_r}$$

The expected probability of a type *t* with sample frequency *f(t) = r* is estimated by *r\*/N*, where *N* is the total number of observed types. Good-Turing obtains good estimates for *r\*/N* if $N_r$ is large. We will see that for our applications, $N_r$ tends to be large for small frequencies *r*, while on the other hand, if $N_r$ is small, *r* is usually large and needs not to be adjusted.

For the adjustment of the frequencies of unseen types, where *r = 0*, *r\** is equal to $N_1/N_0$, where $N_0$ is the number of types that we have not seen. $N_0$ is equal to the difference between the total number of types and the number of observed types. Thus, in order to calculate the adjusted frequency of an unseen type, one needs to know the total number of types in the population. Notice that Good-Turing does not differentiate among the types that have not been seen: the adjusted frequencies of all unseen types are identical.

### 4.3.3   Using Good-Turing to adjust the frequencies of subtrees

The use of the Good-Turing method in natural language technology is far from new. It is commonly applied in speech recognition and part-of-speech tagging for adjusting the frequencies of (un)seen word sequences (e.g. Jelinek, 1985; Katz, 1987; Church & Gale, 1991). In stochastic parsing, Good-Turing has to our knowledge never been tried out. Designers of stochastic parsers seem to have given up on the problem of creating a statistically adequate theory concerning parsing unknown events. Stochastic parsing systems either use a closed lexicon, or use a two step approach where first the words are tagged by a stochastic tagger, after which the p-o-s tags (with or without the words) are parsed by a stochastic parser. The latter approach has become increasingly popular (e.g. Schabes et al., 1993; Weischedel et al., 1993; Briscoe, 1994; Magerman, 1995; Collins, 1996). Notice, however, that the tagger used in this two step approach often uses Good-Turing (or a similar smoothing method) to adjust the observed frequencies of $n$-grams. So why not apply Good-Turing directly to the structural units of a stochastic grammar?

This lack of interest in using Good-Turing may be due to the fact that many stochastic grammars are still being constructed within the *grammar-building* community. In this community, it is generally assumed that grammars need to be as succinct as possible. The existence of unobserved rules is unacceptable from such a *competence* point of view.[4] But from a *performance* point of view, it is very well acceptable that not all statistical units (in our case, subtrees) have been seen; therefore we will put forward the Good-Turing estimator as a statistically and cognitively adequate extension of DOP1.

How can Good-Turing be used for adjusting the frequencies of known and unknown subtrees? It may be evident that it is too rough to apply Good-Turing to all subtrees together. We must distinguish between subtrees of different roots, since in DOP, the spaces of subtrees of a certain root constitute different distributions, for each of which the substitution-probabilities sum up to one. Therefore, Good-Turing is applied to each subtree-class separately, that is, to the *S*-subtrees, *NP*-subtrees, *VP*-subtrees, *N*-subtrees, *V*-subtrees, etc. As in the previous section, we will take only the subtrees upto depth three.

In order to clarify this, we show in table 4.2 the adjusted frequencies for a class of 118348 *NP*-subtrees. The first column of the table shows the observed frequencies of *NP*-subtrees from zero to six. The second column shows $N_r$, the number of *NP*-subtrees that had those frequencies in the training set (the estimation of $N_0$ is a special case and will be dealt with shortly). The third column shows the adjusted frequencies as calculated by the Good-Turing formula.

| $r$ | $N_r$ | $r^*$ |
|---|---|---|
| 0 | 1100000000 | 0.000055 |
| 1 | 60416 | 0.30 |
| 2 | 9057 | 1.37 |
| 3 | 4161 | 1.86 |
| 4 | 1944 | 1.99 |
| 5 | 773 | 3.74 |
| 6 | 482 | 4.37 |

Table 4.2. Adjusted frequencies for *NP*-subtrees

The calculations for $r = 0$ rest on an estimation of $N_0$, the number of *NP*-subtrees that have not been seen. $N_0$ is equal to the difference between the total number of distinct *NP*-subtrees and the number of distinct *NP*-subtrees seen. Thus, we must estimate the total number of possible *NP*-subtrees. To make such an estimation feasible, we use the following assumptions:

* No subtree is larger than depth 3. This was already assumed.
* The unknownness of unseen subtrees only depends on the terminals. Also this was assumed before. It implies that all unlexicalized *NP*-subtrees (i.e. all *NP*-subtrees without words) are known.

---

[4] Although the opposite opinion may be heard as well (e.g. Sampson, 1987).

* The *size* of the vocabulary is known. This is common practice in corpus linguistics, where the estimations are usually restricted to the domain under study. For instance, for the estimation of the population of bigrams the number of distinct unigrams is usually assumed to be known. In our case, we happen to know that the whole ATIS domain contains 1508 distinct words.

Our calculation of (1) the total number of distinct *NP*-subtrees, and (2) $N_0$, can now be accomplished as follows:
(1) The total number of *NP*-subtrees (that can be the result of the mismatch-method) is calculated by attaching in all possible ways 1508 dummies to the tags of the unlexicalized *NP*-subtrees from the training set. This yields a number of $1.10259 \times 10^9$ distinct subtrees. To this number, the number of distinct unlexicalized *NP*-subtrees must be added (12429), yielding $1.10260 \times 10^9$ types for the total number of distinct *NP*-subtrees.
(2) The number of unseen types $N_0$ is the difference between the total number of distinct *NP*-subtrees and the observed number of distinct *NP*-subtrees, $\Sigma_{r>0} N_r$, which is $1.10260 \times 10^9$ - 77479 = $1.10253 \times 10^9$.

Coming back to our adjustment of the frequency of unseen *NP*-subtrees, this can now be calculated by Good-Turing as $N_1/N_0 = 60416/1.1 \times 10^9 \approx 0.000055$.

### 4.4 The model DOP3

DOP3 is very similar to DOP1. What is different in DOP3 is (1) a much larger space of subtrees, which is extended to include subtrees in which one or more terminals are treated as wildcards, and (2) the frequencies of the subtrees, that are now adjusted by the Good-Turing estimator. The probability definitions of derivation and parse in DOP3 are the same as in DOP1.

As to the computational aspects, we can very easily extend the parsing algorithms designed for DOP1 to DOP3, by allowing the terminals of subtrees to mismatch with the words of the input sentence. After assigning the adjusted probabilities to the subtrees in the resulting parse forest, the most probable parse can be estimated in the same way as in DOP1 by Monte Carlo.

In (Bod, 1995a), some interesting properties of DOP3 are derived. Among others, it is shown that DOP3 displays a preference for parses constructed by generalizing over a minimal number of words, and that DOP3 prefers parses that generalize over open-class words to parses that generalize over closed-class words.

### 4.5 Experimental aspects of DOP3

Treating all words as potential unknown-category words would certainly lead to an impractically large number of subtrees in the parse forest. As we have seen, the set of possible *NP*-subtrees (of maximal depth three) consists of $10^9$ types, which is a factor 12000 larger than the set of seen *NP*-subtrees ($8 \times 10^4$). It is therefore evident that we will get impractical processing times with DOP3.

If we still want to perform experiments with DOP3, we need to limit the mismatches as much as possible. It seems reasonable to allow the mismatches only for unknown words, and for a restricted set of potential unknown-category words. From the ATIS training set we derive that only nouns and verbs are actually lexically ambiguous. In our experiments, we will therefore limit the potential unknown-category words of an input sentence to the nouns and verbs. This means that only the words which are unknown in the training set and the words of the test sentence which are tagged as a noun or a verb in the training set are allowed to mismatch with subtree-terminals.

We used the initial random division of the ATIS corpus into a training set of 500 trees and a test set of 100 trees. In order to carefully study the experimental merits of DOP3, we distinguished two classes of test sentences:

1. test sentences containing both unknown and unknown-category words
2. test sentences containing only unknown-category words

Note that all 100 test sentences contained at least one potential unknown-category word (verb or noun). The following table shows the parse accuracy for subtree-depth ≤ 3.

| test sentences | parse accuracy |
|---|---|
| with unknown words and unknown-category words | 34% |
| with only unknown-category words | 57% |
| all test sentences | 41% |

Table 4.3. Parse accuracy for word strings from the ATIS corpus by DOP3

The table shows that DOP3 has better performance than DOP2 in all respects (cf. table 4.1). The parse accuracy for the sentences with unknown and unknown-category words is with 34% much higher than the 20% of DOP2. As to the sentences with only unknown-category words, the improvement of DOP3 with respect to DOP2 is most noticeable: the accuracy increased from 33% to 57%. However, the comparison with DOP2 may not be fair, as DOP2 cannot deal with unknown-category words at all. What the parse results of DOP3 do indicate is, that, for sentences without unknown words, the parse accuracy for word strings is of the same order as the parse accuracy for p-o-s strings (which was 56% at maximum depth 3; see section 4.2). Nevertheless, the total parse accuracy of 41% is still bad.

### 4.6     Enriching DOP3 with a dictionary: the model DOP4

A method which may further improve the accuracy of DOP3 may lie in the use of an external dictionary. In the absence of morphological annotations in ATIS, a dictionary can provide the lexical categories of both known and unknown words of an input sentence. Unfortunately, the ATIS corpus contains several abbreviations and proper nouns that are not found in a dictionary, and which therefore still need to be treated as unknown by means of DOP3. In the following, we will refer to the extension of DOP3 with an external dictionary as DOP4. DOP4 puts all lexical categories (p-o-s tags) of the sentence words, as found in a dictionary, in the chart. Secondly, the sentence is parsed by DOP3, provided that subtree-terminals are allowed to mismatch only with the words that were not found in the dictionary.

In our experiments, we used Longman Dictionary (Longman, 1988) to assign lexical categories to the words of the test sentences. The lexical categories used in Longman are not equal to the lexical categories used in the ATIS corpus, and needed to be converted. The following table shows the results of DOP4 compared with those of DOP3.

| test sentences | parse accuracy | |
|---|---|---|
| | DOP3 | DOP4 |
| with unknown words and unknown-category words | 34% | 47% |
| with only unknown-category words | 57% | 60% |
| all test sentences | 41% | 51% |

Table 4.4. Parse accuracy for word strings from the ATIS corpus by DOP4 against DOP3.

The table shows that there is a considerable increase in parse accuracy from 34% to 47% for sentences with unknown words, while the accuracy for sentences with only unknown-category words shows a slight increase from 57% to 60%. The total parse accuracy of DOP4 reaches 51%. The following table also shows the corresponding *sentence accuracy* and *bracketing accuracy* of DOP4 (for all test sentences together).

| Accuracy metric | DOP4 |
|---|---|
| Parse accuracy | 51% |
| Sentence accuracy | 68% |
| Bracketing accuracy | 93.2% |

Table 4.5. Accuracies for ATIS word strings by DOP4

The above accuracies are the best published results on Penn Treebank ATIS word strings, to the best of our knowledge. Moreover, our results are still based on a limited version of DOP, since for time cost reasons no subtrees larger than depth 3 were used.

We must keep in mind that DOP4 is a hybrid model, where frequencies of subtrees are combined with a dictionary look-up. What we hope to have shown, is, that it is possible to extend a stochastic parsing model in a statistically and cognitively adequate way, such that it can directly parse and disambiguate word strings that contain unknown (-category) words without the need of an external part-of-speech tagger.

# Conclusion

In this paper we have addressed two, previously neglected questions about the DOP model: how does DOP perform if tested on *unedited* Penn Treebank data, and (2), how can DOP be used for directly parsing *word strings* that contain unknown words. We have shown that although parse results are considerably lower on unedited data than on cleaned-up data, they are very competitive, if not better than other models. With respect to the parsing of word strings, we have shown that the hardness of the problem does not lie so much in unknown words, but in previously unseen lexical categories of known words. We have given a novel method for parsing these words by estimating the probabilities of unknown subtrees. The method was tested on ATIS trees obtaining results that to the best of our knowledge are not exceeded by other stochastic parsers. Moreover, the results of a less-than-optimal version of DOP on the Wall Street Journal corpus suggest that the approach can be succesfully extended to larger domains. As future research, we will apply the full DOP model on WSJ word strings in order to compare our results with the best known parsers on this domain (Magerman, 1995; Collins, 1996).

# Acknowledgements

I am grateful to Remko Scha for many useful comments and additions. I also thank three anonymous reviewers for their comments.

# References

J. Baker, 1979. "Trainable grammars for speech recognition", In J. Wolf & D. Klatt (eds.) *Speech communication papers presented at the 97th Meeting of the Acoustical Society of America*, MIT, Cambridge, MA.

R. Bod, 1992. "Data Oriented Parsing (DOP)", *Proceedings COLING'92*, Nantes.

R. Bod, 1993a. "Using an Annotated Corpus as a Stochastic Grammar", *Proceedings EACL'93*, Utrecht.

R. Bod, 1993b. "Monte Carlo Parsing", *Proceedings Third International Workshop on Parsing Technologies*, Tilburg/Durbuy.

R. Bod, 1995a. *Enriching Linguistics with Statistics: Performance Models of Natural Language*, ILLC Dissertation Series 1995-14, University of Amsterdam (obtainable via anonymous ftp: ftp://ftp.fwi.uva.nl/pub/theory/illc/dissertations/DS-95-14.text.ps.gz).

R. Bod, 1995b. "Monte Carlo Parsing", in H. Bunt and M. Tomita (eds.) *Recent Advances in Parsing Technology*, Kluwer Academic Publishers.

R. Bod, 1996. "Efficient Algorithms for Parsing the DOP Model? A Reply to Joshua Goodman." *CMP-LG/9605031*.

R. Bod, R. Bonnema and R. Scha, 1996. "A Data-Oriented Approach to Semantic Interpretation", to appear in *Proceedings Workshop on Corpus-Oriented Semantic Analysis*, ECAI-96, Budapest. (also in *CMP-LG*)

T. Briscoe, 1994. "Prospects for Practical Parsing of Unrestricted Text: Robust Statistical Parsing Techniques", N. Oostdijk and P. de Haan (eds.), *Corpus-based Research into Language*, Rodopi, Amsterdam.

E. Brill, 1993. "Transformation-Based Error-Driven Parsing", *Proceedings Third International Workshop on Parsing Technologies*, Tilburg/Durbuy.

D. Carter and M. Rayner, 1994. "The Speech-Language Interface in the Spoken Language Translator", L. Boves & A. Nijholt (eds.) *Speech and Language Engineering, Proceedings of the eighth Twente Workshop on Language Technology*, Enschede.

M. Collins, 1996. "A New Statistical Parser Based on Bigram Lexical Dependencies", to appear in *Proceedings ACL-96*, Santa Cruz (CA).

E. Charniak, 1996. "Tree-bank Grammars", *Technical Report CS-96-02*, Department of Computer Science, Brown University. (obtainable via WWW home page of Eugene Charniak)

K. Church, 1988. "A Stochastic Parts Program and Noun Phrase Parser for Unrestricted Text", *Proceedings ANLP'88*, Austin, Texas.

K. Church and W. Gale, 1991. "A comparison of the enhanced Good-Turing and deleted estimation methods for estimating probabilities of English bigrams", *Computer Speech and Language* 5, 19-54.

I. Good, 1953. "The Population Frequencies of Species and the Estimation of Population Parameters", *Biometrika* 40, 237-264.

J. Goodman, 1996. "Efficient Algorithms for Parsing the DOP Model", *Proceedings Empirical Methods in Natural Language Processing*, Philadelphia.

F. Jelinek, 1985. "The Development of an Experimental Discrete Dictation Recognizer", *IEEE'85* (Invited Paper).

F. Jelinek and R. Mercer, 1985. "Probability Distribution Estimation from Sparse Data", *IBM Technical Disclosure Bulletin* 28, 2591-2594.

S. Katz, 1987. "Estimation of probabilities from sparse data for the language model component of a speech recognizer", *IEEE Transactions on Acoustics, Speech, and Signal Processing*, ASSP-35, 400-401.

Longman Dictionary of the English Language, 1988, Longman, London.

D. Magerman, 1995. "Statistical Decision-Tree Models for Parsing", *Proceedings ACL'95*, Cambridge, Massachusetts.

C. de Marcken, 1995. "Lexical Heads, Phrase Structure and the Induction of Grammar", *Proceedings of the Third Workshop on Very Large Corpora* (WVLC-3), Cambridge (Mass.).

M. Marcus, B. Santorini and M. Marcinkiewicz, 1993. "Building a Large Annotated Corpus of English: the Penn Treebank", *Computational Linguistics* 19(2).

F. Pereira and Y. Schabes, 1992. "Inside-Outside Reestimation from Partially Bracketed Corpora", *Proceedings ACL'92*, Newark.

M. Rajman, 1995a. "Approche Probabiliste de l'Analyse Syntaxique", *Traitement Automatique des Langues*, vol. 36(1-2).

M. Rajman, 1995b. *Apports d'une approche a base de corpus aux techniques de traitement automatique du langage naturel*, PhD thesis, Ecole Nationale Superieure des Telecommunications, Paris.

A. Ratnaparkhi, 1996. "A Maximum Entropy Model for Part-Of-Speech Tagging", *Proceedings Empirical Methods in Natural Language Processing*, Philadelphia.

G. Sampson, 1987. "Evidence against the 'Grammatical/Ungrammatical' Distinction", W. Meijs (ed.), *Corpus Linguistics and Beyond*, Rodopi, Amsterdam.

R. Scha, 1990. "Taaltheorie en Taaltechnologie; Competence en Performance", in Q.A.M. de Kort and G.L.J. Leerdam (eds.), *Computertoepassingen in de Neerlandistiek*, Almere: Landelijke Vereniging van Neerlandici (LVVN-jaarboek).

Y. Schabes, M. Roth and R. Osborne, 1993. "Parsing the Wall Street Journal with the Inside-Outside Algorithm", *Proceedings EACL'93*, Utrecht.

S. Sekine and R. Grishman, 1995. "A Corpus-based Probabilistic Grammar with Only Two Non-terminals", *Proceedings Fourth International Workshop on Parsing Technologies*, Prague.

K. Sima'an, 1995. "An optimized algorithm for Data Oriented Parsing", *Proceedings International Conference on Recent Advances in Natural Language Processing*, Tzigov Chark, Bulgaria.

K. Sima'an, 1996. "An optimized algorithm for Data Oriented Parsing", in R. Mitkov and N. Nicolov (eds.), *Recent Advances in Natural Language Processing 1995*, volume 136 of *Current Issues in Linguistic Theory*. John Benjamins, Amsterdam.

R. Weischedel, M. Meteer, R, Schwarz, L. Ramshaw and J. Palmucci, 1993. "Coping with Ambiguity and Unknown Words through Probabilistic Models", *Computational Linguistics*, 19(2), 359-382.